\pdfoutput=1

\documentclass[11pt]{article}

\usepackage[]{emnlp2021}

\usepackage{times}
\usepackage{latexsym}

\usepackage{amsfonts}
\usepackage{amsmath}
\usepackage{bm}
\usepackage{tabularx}
\usepackage{url}
\usepackage{multirow}
\usepackage{hyperref}
\usepackage{booktabs}
\usepackage{graphicx}
\usepackage{booktabs}
\usepackage{floatrow}

\newcommand{\bert}{\textsc{Bert}}
\newcommand{\roberta}{\textsc{RoBERTa}}
\newcommand{\electra}{\textsc{Electra}}
\newcommand{\albert}{\textsc{Albert}}
\newcommand{\glue}{\textsc{Glue}}
\newcommand{\squad}{\textsc{SQuAD}}
\newcommand{\mlm}{\textsc{mlm}}
\newcommand{\nsp}{\textsc{nsp}}

\usepackage[T1]{fontenc}

\usepackage[utf8]{inputenc}

\usepackage{microtype}

\title{Frustratingly Simple Pretraining Alternatives \\to Masked Language Modeling}

\author{
Atsuki Yamaguchi{\rm \textsuperscript{1}}\Thanks{ Work was done while at the University of Sheffield.} , George Chrysostomou{\rm \textsuperscript{2}}, Katerina Margatina{\rm \textsuperscript{2}} {\rm and} Nikolaos Aletras{\rm \textsuperscript{2}} \\
\textsuperscript{1}{Research and Development Group, Hitachi,  Ltd., Japan} \\
\textsuperscript{2}{Department of Computer Science, University of Sheffield, United Kingdom}\\
\textsuperscript{1}\texttt{atsuki.yamaguchi1@gmail.com}\\
\textsuperscript{2}\texttt{\{gchrysostomou1, k.margatina, n.aletras\}@sheffield.ac.uk}\\
}

\begin{document}
\maketitle
\begin{abstract}
Masked language modeling (\mlm{}), a self-supervised pretraining objective, is widely used in natural language processing for learning text representations. \mlm{} trains a model to predict a random sample of input tokens that have been replaced by a \texttt{[MASK]} placeholder in a multi-class setting over the entire vocabulary.
When pretraining, it is common to use alongside \mlm{} other auxiliary objectives on the token or sequence level to improve downstream performance (e.g. next sentence prediction). However, no previous work so far has attempted in examining whether other simpler linguistically intuitive or not objectives can be used standalone as main pretraining objectives.
In this paper, we explore five simple pretraining objectives based on token-level classification tasks as replacements of \mlm{}.  
Empirical results on \glue{} and \squad{} show that our proposed methods achieve comparable or better performance to \mlm{} using a \bert{}-\textsc{base} architecture. We further validate our methods using smaller models,
showing that pretraining a model with 41\% of the \bert{}-\textsc{base}'s parameters, \bert{}-\textsc{medium} results in only a 1\% drop in \glue{} scores with our best objective.\footnote{Our code is publicly available here: \url{https://github.com/gucci-j/light-transformer-emnlp2021}}

\end{abstract}

\section{Introduction}\label{sec:intro}

Masked Language Modeling (\mlm{}) pretraining~\citep{Devlin2019, Liu2019, Lan2020ALBERT:, Wang2020StructBERT:} is widely used in natural language processing (NLP) for self-supervised learning of text representations. \mlm{} trains a model (typically a neural network) to predict a particular token that has been replaced with a \texttt{[MASK]} placeholder given its surrounding context. \citet{Devlin2019} first proposed \mlm{} with an additional next sentence prediction (\nsp{}) task (i.e. predicting whether two segments appear consecutively in the original text) to train \bert{}. 

Recently several studies have extended \mlm{}, by masking a contiguous segment of the input instead of treating each token independently \citep{song2019mass, Sun_Wang_Li_Feng_Tian_Wu_Wang_2020, joshi-etal-2020-spanbert}.
\citet{NEURIPS2019_dc6a7e65} reformulated \mlm{} in \textsc{Xlnet}, to mask out attention weights rather than input tokens, such that the input sequence is auto-regressively generated in a random order. 
\textsc{Electric} \citep{clark-etal-2020-pre} addressed the expensive softmax issue of \mlm{} using a binary classification task, where the task is to distinguish between words sampled from the original data distribution and a noise distribution, using noise-contrastive estimation.
In a different direction, previous work has also developed methods to complement \mlm{} for improving text representation learning.
\citet{Aroca2020} have explored sentence and token-level auxiliary pretraining objectives, showing improvements over \nsp{}. \albert{}~\cite{Lan2020ALBERT:} complemented \mlm{} with a similar task that predicts whether two sentences are in correct order or swapped. \electra{}~\cite{Clark2020ELECTRA:} introduced a two-stage token-level prediction task; using a \mlm{} generator to replace input tokens and subsequently a discriminator trying to predict whether a token has been replaced or not.

Despite these advances, simpler linguistically motivated or not auxiliary objective tasks acting as primary pre-training objectives substituting completely \mlm{} have not been explored. Motivated by this, we propose five frustratingly simple pretraining tasks, showing that they result into models that perform competitively to \mlm{} when pretrained for the same duration (e.g. five days) and fine-tuned in downstream tasks in \glue{} \cite{Wang2018} and \squad{} \cite{Rajpurkar2016} benchmarks. 

\begin{figure*}[t!]
    \begin{center}
    \includegraphics[width=1\linewidth]{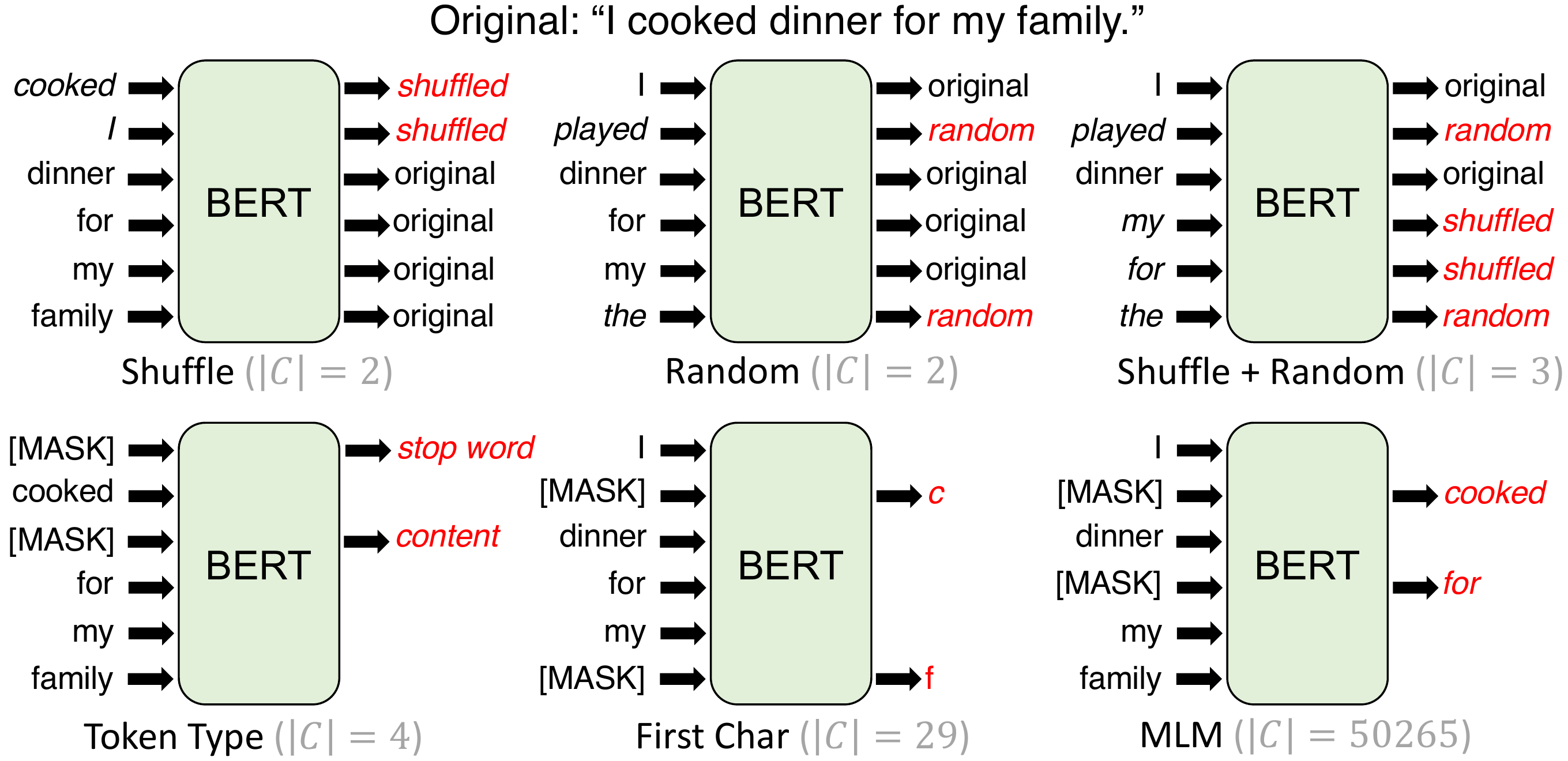}
    \caption{Overview of our five frustratingly simple pretraining tasks along with a comparison to \mlm{}. $|C|$ denotes the number of classes for each task.}
    \label{fig:tasks}
    \end{center}
\end{figure*}

\noindent\textbf{Contributions: } (1) To the best of our knowledge, this study is the first to investigate  whether linguistically and non-linguistically intuitive tasks can effectively be used for pretraining (\S\ref{sec:pretraining_obj}). (2) We empirically demonstrate that our proposed objectives are often computationally cheaper and result in better or comparable performance to \mlm{} across different sized models (\S\ref{sec:results}).

\section{Pretraining Tasks}\label{sec:pretraining_obj}

Our methodology is based on two main hypotheses: (1) effective pretraining should be possible with standalone token-level prediction methods that are linguistically intuitive (e.g. predicting whether a token has been shuffled or not should help a model to learn semantic and syntactic relations between words in a sequence); and (2) the deep architecture of transformer models should allow them to learn associations between input tokens even if the pretraining objective is not linguistically intuitive (e.g. predicting the first character of a masked token should not matter for the model to learn that `cat' and `sat' usually appear in the same context). Figure \ref{fig:tasks} illustrates our five linguistically and non-linguistically intuitive pretraining tasks with a comparison to \mlm{}.

\paragraph{Shuffled Word Detection (\textsc{Shuffle}):} Motivated by the success of \electra{}, our first pretraining objective is a token-level binary classification task, consisting of identifying whether a token in the input sequence has been shuffled or not. For each sample, we randomly shuffle 15\% of the tokens. This task is trained with the \textit{token-level} binary cross-entropy loss averaged over all input tokens (i.e. shuffled and original). The major difference between ours and \electra{} is that we do not rely on \mlm{} to replace tokens. Our intuition is that a model can acquire both syntactic and semantic knowledge by distinguishing shuffled tokens in context.

\paragraph{Random Word Detection (\textsc{Random}):}

We now consider replacing tokens with out-of-sequence tokens. For this purpose we propose \textsc{Random}, a pretraining objective which replaces 15\% of tokens with random ones from the vocabulary. Similar to shuffling tokens in the input, we expect that replacing a token in the input with a random word from the vocabulary ``forces'' the model to acquire both syntactic and semantic knowledge from the context to base its decision on whether it has been replaced or not.

\paragraph{Manipulated Word Detection (\textsc{Shuffle + Random}):} 
For our third pretraining objective, we seek to increase the task difficulty and subsequently aim to improve the text representations learned by the model. We therefore propose an extension of \textsc{Shuffle} and \textsc{Random}, which is a three-way token-level classification task for predicting whether a token is a shuffled token, a random token, or an original token. For each sample, we replace 10\% of tokens with shuffled ones from the same sequence and another 10\% of tokens with random ones from the vocabulary. This task can be considered as a more complex one, because the model must recognize the difference between tokens replaced in the same context and tokens replaced outside of the context. For this task we use the cross-entropy loss averaged over all input tokens.

\paragraph{Masked Token Type Classification (\textsc{Token Type}):}
Our fourth objective is a four-way classification, aiming to predict whether a token is a stop word,\footnote{We use the Natural Language Toolkit's stop word list: \url{https://www.nltk.org/}.} a digit, a punctuation mark, or a content word. Therefore, the task can be seen as a simplified version of POS tagging. We regard any tokens that are not included in the first three categories as content words. We mask 15\% of tokens in each sample with a special \texttt{[MASK]} token and compute the cross-entropy loss over the masked ones only not to make the task trivial. For example, if we compute the token-level loss over unmasked tokens, a model can easily recognize the four categories as we only have a small number of non-content words in the vocabulary.

\paragraph{Masked First Character Prediction (\textsc{First Char}):}
Finally to test our second hypothesis, we propose a simplified version of the \mlm{} task, where the model has to predict only the first character of each masked token instead of performing a softmax over the entire vocabulary. We define a 29-way classification task, where 29 categories include the English alphabet (0 to 25), a digit (26), a punctuation mark (27), or any other character (28). We mask 15\% of tokens in each sample and compute the cross-entropy loss over the masked tokens only.\footnote{For more details on the pretraining tasks, including equations, see Appendix~\ref{appendix:task}.}

\section{Experimental Setup}\label{sec:exp}

\paragraph{Models:}\label{sec:model_arch} 
We use \bert{}~\cite{Devlin2019} (\textsc{base}) as our basis model by replacing the \mlm{} and \nsp{} objectives with one of our five token-level pretraining tasks in all our experiments. We also consider two smaller models from \citet{Turc2019wellread}, \textsc{medium}  and \textsc{small}, where we reduce the size of the following components compared to the \textsc{base} model: (1) hidden layers; (2) hidden size; (3) feed-forward layer size; and (4) attention heads. More specifically, \textsc{medium} has eight hidden layers and attention heads, while \textsc{small} has four hidden layers and eight attention heads. The size of feed-forward and hidden layers for both models are 2048 and 512, respectively.

\paragraph{Pretraining Data:}\label{sec:data}

We pretrain all models on the English Wikipedia and BookCorpus \cite{Zhu2015} (WikiBooks) using the \texttt{datasets} library.\footnote{\url{https://github.com/huggingface/datasets}}

\paragraph{Implementation Details:}\label{sec:impl_details}

We pretrain and fine-tune our models with two NVIDIA Tesla V100 (SXM2 - 32GB) with a batch size of 32 for \textsc{base} and 64 for \textsc{medium} and \textsc{small}. We pretrain all our models for up to five days each due to limited access to computational resources and funds for running experiments. We save a checkpoint of each model every 24 hours.\footnote{For more details on model setup, implementation, and data preprocessing, see Appendix \ref{appendix:experimental_setup}.}

\paragraph{Evaluation:}\label{sec:metrics}

We evaluate our approaches on \glue{} \cite{Wang2018} and \squad{} \cite{Rajpurkar2016} benchmarks. To measure performance in downstream tasks, we fine-tune all models for five times each with a different random seed.

\paragraph{Baseline:}\label{sec:baseline}

For comparison, we also pretrain models with \mlm{}. Following \bert{} and \roberta{}, we mask 15\% of tokens in each training instance, where 80\% of the tokens are replaced with \texttt{[MASK]}, 10\% of the tokens are replaced with a random word and the rest of tokens remain unchanged. We compute the cross-entropy loss averaged over the masked tokens only.

\renewcommand*{\arraystretch}{1.0}
\begin{table*}[!t]
\begin{center}
\small
\resizebox{\linewidth}{!}{%
\begin{tabular}{lcccccccc|c|c}
\toprule
Pretraining task & MNLI & QNLI & QQP & RTE & SST & MRPC & CoLA & STS & GLUE Avg. & SQuAD v1.1 \\ \midrule

 & \multicolumn{9}{c}{\textsc{base} - 40 Epochs Pretraining (Upper Bound)} \\ \cmidrule{2-11}
\enskip \mlm{} + \nsp{} & 83.8 & 90.8 & 87.8 & 69.9 & 91.9 & 85.0 & 58.9 & 89.3 & 82.1 (0.4) & 87.4 (0.6) \\ \midrule

Ours & \multicolumn{10}{c}{\textsc{base} - Five Days Pretraining} \\ \cmidrule{2-11}
\enskip \mlm{} & \textbf{80.1} & \textbf{88.2} & 85.9 & 61.4 & \textbf{89.6} & 81.6 & 49.6 & 84.7 & 77.6 (0.2)  & \textbf{84.8 (0.2)} \\

\enskip Shuffle & 73.3 & 81.6 & 82.1 & 57.5 & 82.4 & 79.1 & 33.4 & 79.9 & 71.2 (0.3)  & 74.8 (0.2) \\

\enskip Random & 78.6 & 87.0 & 85.5 & 60.5 & 87.4 & 81.6 & 47.0 & 84.0 & 76.4 (0.2)  & 81.6 (0.4) \\

\enskip Shuffle + Random & 78.6 & 87.7 & \textbf{86.1} & \textbf{65.1} & 87.8 & \textbf{87.0} & \textbf{54.9} & \textbf{86.7} & \textbf{79.2 (0.3)}  & 83.5 (0.2) \\

\enskip Token Type & 75.1 & 84.2 & 83.9 & 56.8 & 86.7 & 75.5 & 40.3 & 77.4 & 72.5 (0.2)  & 78.6 (0.7) \\

\enskip First Char & 78.2 & 87.1 & 85.5 & 60.7 & 89.5 & 83.6 & 43.9 & 84.6 & 76.7 (0.5)  & 82.0 (0.1) \\ \midrule

 & \multicolumn{10}{c}{\textsc{medium} - Five Days Pretraining} \\ \cmidrule{2-11}
\enskip \mlm{} & \textbf{78.7} & 85.3 & 85.4 & 61.7 & \textbf{89.9} & 80.6 & 43.1 & 84.5 & 76.1 (0.4)  & \textbf{81.8 (0.5)} \\

\enskip Shuffle & 77.3 & 86.4 & 85.3 & 64.0 & 87.9 & 83.4 & \textbf{53.8} & 84.1 & 77.8 (0.2)  & 81.3 (0.2) \\

\enskip Random & 77.7 & 86.2 & 85.6 & \textbf{64.3} & 87.8 & 81.7 & 44.3 & 84.8 & 76.6 (0.3)  & 79.5 (0.1) \\

\enskip Shuffle + Random & 78.3 & \textbf{87.0} & \textbf{85.7} & 63.3 & 87.8 & \textbf{85.9} & 52.4 & \textbf{85.4} & \textbf{78.2 (0.2)}  & \textbf{81.8 (0.2)} \\

\enskip Token Type & 76.0 & 84.7 & 84.4 & 59.7 & 87.6 & 81.4 & 45.8 & 80.7 & 75.0 (0.4)  & 79.8 (0.4) \\

\enskip First Char & 77.4 & 85.6 &	85.1 &	59.4 &	88.8 & 83.9 &	42.4 &	83.0 & 75.7 (0.3)  & 79.5 (0.2) \\ \midrule

& \multicolumn{10}{c}{\textsc{small} - Five Days Pretraining} \\ \cmidrule{2-11}
\enskip \mlm{} & 76.2 & 84.2 & 84.8 & 57.5 & \textbf{88.6} & \textbf{82.9} & 36.3 & 83.0 & 74.2 (0.4)  & 77.1 (0.3) \\

\enskip Shuffle & 74.9	& 84.0	& 84.2	& 59.8	& 86.4	& 80.0	& \textbf{47.1} & 81.1 & 74.7 (0.3)  & 76.1 (0.6) \\

\enskip Random & 75.6 & 84.7 & 84.8 & 58.3 & 86.7 & 80.0 & 39.6 & 83.5 & 74.1 (0.4)  & 76.7 (0.5) \\

\enskip Shuffle + Random & \textbf{76.9}	& \textbf{85.7} & \textbf{85.3}	& \textbf{60.3}	& 87.1	& 81.8	& 41.7	& \textbf{84.6} & \textbf{75.4 (0.4)}  & \textbf{77.5 (0.3)} \\

\enskip Token Type & 73.2	& 83.0	& 83.7	& 58.8	& 86.4	& 77.1	& 37.1	& 77.8 & 72.1 (0.4)  & 74.2 (0.3) \\

\enskip First Char & 75.3 & 84.0 & 84.9 & 55.6 & 87.2 & 79.8 & 33.1 & 83.3 & 72.9 (0.8)  & 77.4 (0.2) \\

\bottomrule
\end{tabular}%
}
\caption{Results on GLUE and SQuAD dev sets with standard deviations over five runs in parentheses. For MNLI, we report matched accuracy, for CoLA Matthews correlation, for STS-B Spearman correlation, for MRPC  accuracy, for QQP and SQuAD F1 scores; accuracy for all other tasks. \textbf{Bold} values denote best performing across each dataset and Avg. for each model setting.} 

\label{table:base_result}
\end{center}
\end{table*}

\section{Results}\label{sec:results}

\paragraph{Performance Comparison:}

Table \ref{table:base_result} presents results on \glue{} and \squad{}, for our five pretraining tasks compared to \mlm{} across all model configurations (\S \ref{sec:exp}).  We also include for reference our replicated downstream performance by fine-tuning \bert{}-\textsc{base} (\mlm{} + \nsp{}) pretrained\footnote{We used an already pretrained model provided by \citet{wolf-etal-2020-transformers}.} for 40 epochs (Upper Bound).

We first observe that our best objective, Shuffle + Random, outperforms \mlm{} on \glue{} Avg. and \squad{} in the majority of model settings (\textsc{base}, \textsc{medium} and \textsc{small}) with five days pretraining. 
For example in \glue{}, we obtain an average of 79.2 using Shuffle + Random with \bert{}-\textsc{base} compared to 77.6 using \mlm{}. This suggests that Shuffle + Random can be a competitive alternative to \mlm{}. 
Although Shuffle + Random does not outperform \mlm{} in \squad{} only with \bert{}-\textsc{base}, it remains competitive (83.5 compared to 84.8). The remainder of our proposed tasks perform well, with First Char and Random being close to \mlm{} across all model configurations confirming our two hypotheses. Finally, Shuffle with \bert{}-\textsc{base} records the lowest performance on \glue{} (71.2 points), but it performs best when combined with Random (i.e. Shuffle + Random).

\begin{figure*}[t!]
    \begin{center}
    \includegraphics[width=1\linewidth]{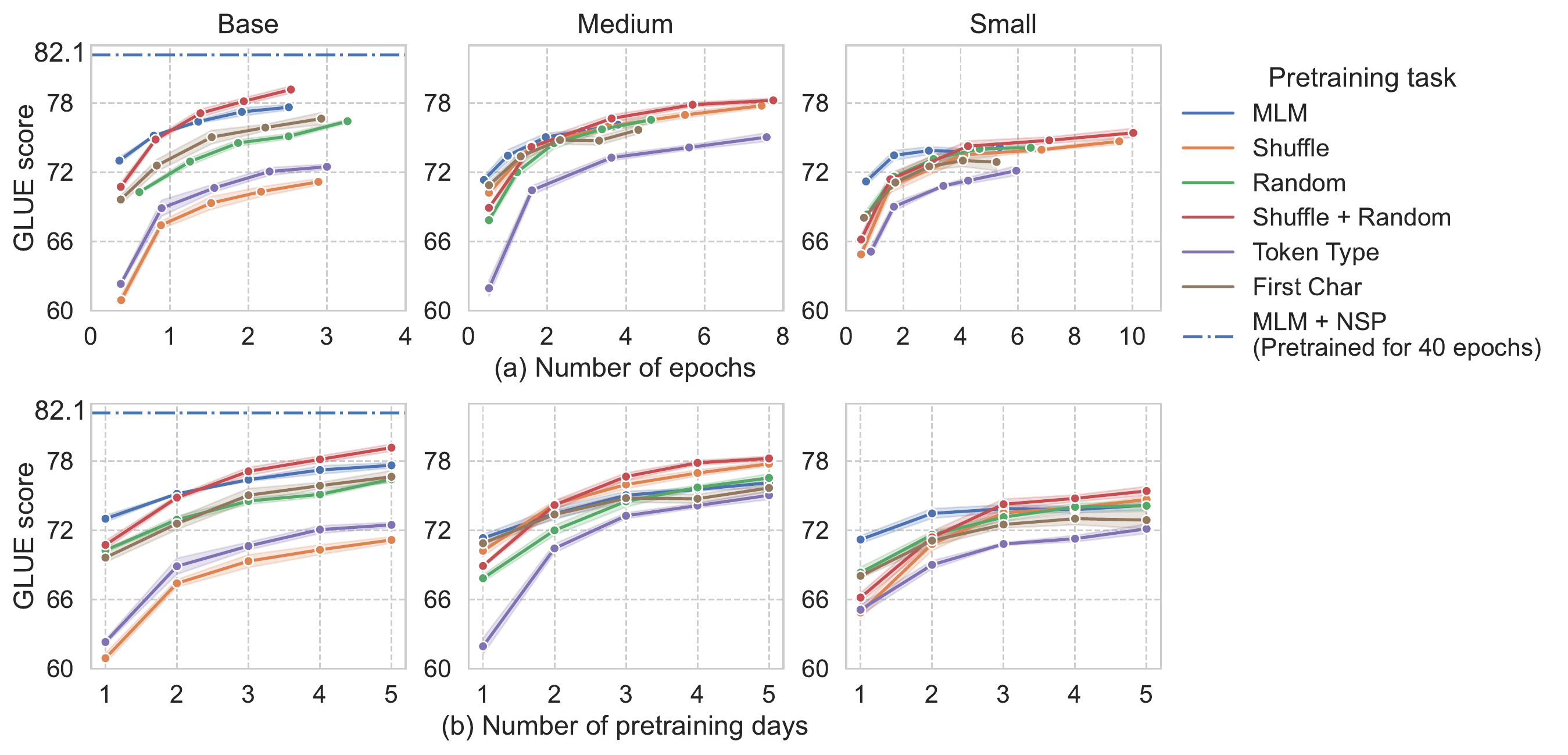}
    \caption{Results on \glue{} dev sets across (a) epochs and (b) days. Each point is a checkpoint pretrained for $1 \leq n \leq 5$ day(s).}
    \label{fig:glue_result}
    \end{center}
\end{figure*}

\paragraph{Computational Efficiency Comparison:}
Figure \ref{fig:glue_result} presents the performance of our proposed methods across (a) epochs and (b) days in \glue{} (\squad{} results available in Appendix~\ref{sec:results_squad_appendix}).
Results suggest that our methods are, in general, more computationally efficient compared to \mlm{}. Shuffle + Random trains for the largest number of epochs (i.e. faster forward-backward passes) in five days for the \textsc{small} and \textsc{medium} settings, with Random outperforming the rest in the \textsc{base} model setting (Figure \ref{fig:glue_result} (a)).
If we take a closer look, we can also see that Shuffle + Random obtains higher performance to \mlm{} across all model configurations when training for a similar number of epochs, suggesting that our approach is a more data efficient task. 
Finally, we can also assume that Shuffle + Random is more challenging than \mlm{} as in all settings it results in lower \glue{} scores after the first day of pretraining (Figure \ref{fig:glue_result} (b)). However, with more iterations it is clear that it results in learning better text representations and quickly outperforms \mlm{}. For example, it achieves a performance of 78.2 compared to 76.1 for \mlm{} with \textsc{medium} on the fifth day. 
Regarding the remainder of our proposed objectives, we can see that they perform comparably and sometimes better than the \mlm{} under \textsc{small} and \textsc{medium} model settings. However, \mlm{} on average outperforms them in the \textsc{base} setting where the models are more highly parameterized.

Lastly, we observe that for the majority of \glue{} tasks, we obtain better or comparable performance to \mlm{} with a maximum of approximately three epochs of training with a \textsc{base} model. This demonstrates that excessively long and computationally inefficient pretraining strategies do not add a lot in downstream performance.

\section{Discussion}
Based on our results, there are mainly two key elements that should be considered for designing pretraining objectives.

\paragraph{Task Difficulty:}
A pretraining task should be moderately difficult to learn in order to induce rich text representations. For example, we can assume from the results that Token Type was somewhat easy for a model to learn as it is a four-way classification of identifying token properties. Besides, in our preliminary experiments, predicting whether a masked token is a stop word or not (Masked Stop Word Detection) also did not exhibit competitive downstream performance to \mlm{} as the task is a lot simpler than Token Type.

\paragraph{Robustness:}
A model should always learn useful representations from ``every'' training sample to solve a pretraining task, regardless of the task difficulty. For instance, Figures \ref{fig:base_loss} to \ref{fig:small_loss} in Appendix \ref{sec:loss_curves} demonstrate that Shuffle needs some time to start converging across all model configurations, which means the model struggled to acquire useful representations at first. In contrast, the loss for Shuffle + Random consistently decreases. Because Shuffle + Random is a multi-class classification, unlike Shuffle or Random, we assume that it can convey richer signals to the model and help stabilize pretraining. Finally, we can also assume that \mlm{} satisfies both elements as it is a multi-class setting over the entire vocabulary and its loss consistently decreases.

\section{Conclusions}

We have proposed five simple self-supervised pretraining objectives and tested their effectiveness against \mlm{} under various model settings. We show that our best performing, manipulated word detection task, results in comparable performance to \mlm{} in \glue{} and \squad{}, whilst also being significantly faster in smaller model settings. We also show that our tasks result in higher performance trained for the same number of epochs as \mlm{}, suggesting higher data efficiency. For future work, we are interested in exploring \textit{which has the most impact in pretraining: the data or the pretraining objective?}

\section*{Acknowledgments}
NA is supported by EPSRC grant EP/V055712/1, part of the European Commission CHIST-ERA programme, call 2019 XAI: Explainable Machine Learning-based Artificial Intelligence. KM is supported by Amazon through the Alexa Fellowship scheme.

\bibliographystyle{emnlp2021_natbib}
\bibliography{anthology,custom}

\appendix
\clearpage 

\section*{Appendices}
\section{Task Details
\label{appendix:task}}
Here, we detail our frustratingly simple pretraining objectives, which are based on token-level classification tasks and can be used on any unlabeled corpora without laborious preprocessing to obtain labels for self-supervision.

\paragraph{Shuffled Word Detection (\textsc{Shuffle}):} Our first pretraining task is a token-level binary classification task, which consists of identifying whether a token in the input sequence has been shuffled or not. For each sample, we randomly shuffle 15\% of the tokens. This task is trained with the \textit{token-level} binary cross-entropy loss averaged over all input tokens:

\begin{equation}
    \begin{split}
    \mathcal{L}_{\mathrm{shuffle}} &= -\frac{1}{N}\sum_{i=1}^{N}y_i\log p(x_i) \\ 
    & + (1 - y_i)\log(1- p(x_i))
    \label{eq:shuffle}
    \end{split}
\end{equation}
\noindent where $N$ is the number of tokens in a sample, and $p(x_i)$ represents the probability of the $i$-th input token $x_i$ predicted as \textit{shuffled} by a model. $y_i$ is the corresponding target label.

This task is motivated by the success of \electra{}, whose pretraining task is to let a discriminator to predict whether a given token is original or replaced (replaced word detection) in addition to \mlm{}. The major difference between ours and \electra{} is that we do not rely on \mlm{}, whereas \electra{} utilizes it as its generator. Here, our intuition is that a model should acquire both syntactic and semantic knowledge to detect shuffled tokens in contexts.

\paragraph{Random Word Detection (\textsc{Random}):} We also consider replacing tokens with out-of-sequence tokens. For this purpose we propose \textsc{Random}, a pretraining objective which replaces 15\% of tokens with random ones from the vocabulary. Similar to shuffling tokens in the input, we expect that replacing a token in the input with a random word from the vocabulary ``forces'' the model to acquire both syntactic and semantic knowledge from the context to base its decision on whether it has been replaced or not. This task is trained with the token-level binary cross-entropy loss averaged over all input tokens (Eq. (\ref{eq:shuffle})).

\paragraph{Manipulated Word Detection (\textsc{Shuffle + Random}):} 
Our third task is a three-way token-level classification of whether a token is a shuffled token, a random token, or an original token. For each sample, we replace 10\% of tokens with shuffled ones and another 10\% of tokens with random ones. This task is an extension of \textsc{Shuffle} and \textsc{Random} and can be regarded as a more complex one because the model must recognize the difference between a token replaced in the same context and a token replaced outside of the context. For this task we employ the cross-entropy loss averaged over all input tokens:

\begin{equation}
    \mathcal{L}_{\mathrm{manipulated}} = -\frac{1}{N}\sum_{i=1}^{N}\sum_{j=1}^{3}y_{ij}\log p_{ij}(x_i)
    \label{eq:manipulated}
\end{equation}
\noindent where $p_{ij}(x_i)$ represents the probability of the $i$-th input token $x_i$ predicted as \textit{shuffled} ($j=1$), \textit{randomized} ($j=2$), or \textit{original} ($j=3$) by a model. $y_{ij}$ is the corresponding target label.

\paragraph{Masked Token Type Classification (\textsc{Token Type}):}
Our fourth task is a four-way classification task that identifies whether a token is a stop word\footnote{A stop word category is based on the Natural Language Toolkit's stop word list: \url{https://www.nltk.org/}.}, a digit, a punctuation mark, or a content word. We regard any tokens that are not included in the first three categories as content words. We mask 15\% of tokens in each sample with a special \texttt{[MASK]} token and compute the cross-entropy loss over the masked ones only not to make the task trivial: if we compute the token-level loss, including unmasked tokens, a model can easily recognize the four categories of tokens as we have a small number of tokens for non-content words.
In this task, a model should be able to identify the distinction between different types of tokens; therefore, the task can be seen as a simplified version of POS tagging.

\paragraph{Masked First Character Prediction (\textsc{First Char}):}
Our last task is a 29-way classification task, where a model needs to predict the first character of a masked token. The 29 categories include the English alphabet (0 to 25), a digit (26), a punctuation mark (27), or any other character (28). We mask 15\% of tokens in each sample and compute the cross-entropy loss over the masked ones only. This task can be seen as a simplified version of \mlm{} as the model just need to predict the first character of each masked token. Besides, it is also similar to masked character-level language modeling, in that the output of both tasks is in characters.

\section{Non-linguistically Intuitive Task}
As we have described in Section \ref{sec:pretraining_obj}, a non-linguistically intuitive task should not be ``explicitly'' related to an input sequence to solve, unlike linguistically intuitive tasks, such as Shuffle and Random. For example, predicting the first character of a masked token should not matter for a model to learn that `cat' and `sat' usually appear in the same context. However, because accurately predicting the first character requires the model to guess its whole word ``implicitly'' given its surrounding tokens, the first character of each masked token should be related to the context. The deep architecture of transformer-based models should allow them to learn such ``implicit'' associations between input tokens by solving the non-linguistically intuitive task, which leads to helping them to learn syntactic and semantic relations between tokens.

\section{Experimental Setup 
\label{appendix:experimental_setup}}

\subsection{Model Architecture}\label{sec:model_arch_appendix}
For all our experiments, we use \bert{}~\cite{Devlin2019} as our basis model by replacing the \mlm{} and \nsp{} objectives with one of our five token-level pretraining tasks. More specifically, we employ \bert{}-\textsc{base} (12 hidden layers and attention heads, $Dim_{\mathrm{hidden}}=768$, $Dim_{\mathrm{intermediate}}=3072$, $\mbox{Total parameters}=125M$) (\textsc{base}), \textsc{medium} (eight hidden layers and attention heads,   $Dim_{\mathrm{hidden}}=512$, $Dim_{\mathrm{intermediate}}=2048$,  $\mbox{Total parameters}=51.5M$), and \textsc{small} (four hidden layers and eight attention heads,   $Dim_{\mathrm{hidden}}=512$, $Dim_{\mathrm{intermediate}}=2048$ $\mbox{Total parameters}=38.9M$).

\subsection{Data}\label{sec:data_appendix}

Following \citet{Devlin2019}, we use the English Wikipedia and BookCorpus \cite{Zhu2015} data (WikiBooks) downloaded from the \texttt{datasets} library\footnote{\url{https://github.com/huggingface/datasets}}. We remove headers for the English Wikipedia and extract training samples with a maximum length of 512. For the BookCorpus, we concatenate sentences such that the total number of tokens is less than 512. For the English Wikipedia, we extract one sample from articles whose length is less than 512. We tokenize text using byte-level Byte-Pair-Encoding \cite{sennrich-etal-2016-neural}. The resulting corpus consists of 8.1 million samples and 2.7 billion tokens in total.

\subsection{Implementation Details}\label{sec:impl_details_appendix}

We implement our models using PyTorch \cite{PyTorch} and the \texttt{transformers} library \cite{wolf-etal-2020-transformers}. We pretrain our models with two NVIDIA Tesla V100 (SXM2 - 32GB) and use one for fine-tuning. Our code is publicly available on GitHub: \url{https://github.com/gucci-j/light-transformer-emnlp2021}.

\paragraph{Pretraining:}
We set the batch size to 32 for the \textsc{base} models and 64 for the \textsc{medium} and \textsc{small} models. We pretrain models for five days and optimized them with an Adam optimizer \cite{Kingma2014}. We apply automatic mixed precision and distributed training during pretraining. Note that we generate labels dynamically during pretraining.

\paragraph{Finetuning:}
We fine-tune models for up to 10 and 20 epochs with early stopping for \squad{} and \glue{}, respectively. To minimize the effect of random seeds, we test five different random seeds for each task. We omitted the problematic WNLI task for \glue{}, following \citet{Aroca2020}.

\subsection{Hyperparameter Details
\label{appendix:hyperparameter}}

As explained in Section \ref{sec:exp}, we entirely followed the \bert{} architecture and only modified its output layer depending on the task employed. Table \ref{tb:params} shows the hyperparameter settings for pretraining and fine-tuning. Note that we utilized neither any parameter sharing tricks nor any techniques that did not appear in \citet{Devlin2019}.

\begin{table*}[th]
\small
\begin{center}
\resizebox{\linewidth}{!}{%
\begin{tabular}{lcc}
\toprule
\textbf{Hyperparameter} & \textbf{Pretraining} & \textbf{Fine-tuning}\\
\midrule
\multicolumn{1}{l}{\multirow{2}{*}{Maximum train epochs}} & \multicolumn{1}{l}{10 epochs for \textsc{base} and \textsc{medium}}
&\multicolumn{1}{l}{Up to 20 epochs for GLUE}\\
&\multicolumn{1}{l}{15 epochs for \textsc{small}}
&\multicolumn{1}{l}{Up to 10 epochs for SQuAD}\\

\multicolumn{1}{l}{\multirow{2}{*}{Batch size (per GPU)}} & \multicolumn{1}{l}{16 for \textsc{base}}
&\multicolumn{1}{l}{32 for GLUE}\\
&\multicolumn{1}{l}{32 for \textsc{half} and \textsc{quarter}}
&\multicolumn{1}{l}{16 for SQuAD}\\

Adam $\epsilon$ & 1e-8\\
Adam $\beta_1$ & 0.9\\
Adam $\beta_2$ & 0.999\\

\multicolumn{1}{l}{\multirow{2}{*}{Sequence length}} & \multicolumn{1}{c}{\multirow{2}{*}{512}}
& \multicolumn{1}{l}{128 for GLUE}\\
&& \multicolumn{1}{l}{384 for SQuAD}\\

\multicolumn{1}{l}{\multirow{4}{*}{Peak learning rate}} & \multicolumn{1}{l}{\textbf{\textsc{base}}: 1e-4 for MLM and Token Type, 1e-5 for Shuffle.} & \multicolumn{1}{c}{\multirow{4}{*}{3e-5}}\\
&\multicolumn{1}{l}{5e-5 for First Char, Random and Shuffle + Random.}\\
&\multicolumn{1}{l}{\textbf{\textsc{medium} \& \textsc{small}}: 1e-4 for MLM, Token Type and First Char.}\\
&\multicolumn{1}{l}{5e-5 for Shuffle, Random and Shuffle + Random.}\\

Warmup steps & 10000 & First 6\% of steps\\
Weight decay & 0.01 & 0\\
Attention Dropout & 0.1\\
Dropout & 0.1\\
\multicolumn{1}{l}{\multirow{2}{*}{Early stopping criterion}} & & \multicolumn{1}{l}{GLUE: No improvements over 5\% of steps.}\\
& & \multicolumn{1}{l}{SQuAD: No improvements over 2.5\% of steps.}\\

\bottomrule
\end{tabular}%
}
\caption{Hyperparameters in our experiments. If not shown, the hyperparameters for fine-tuning are the same as the pretraining ones.}
\label{tb:params}
\end{center}
\end{table*}

\section{Pretraining Behavior} \label{sec:loss_curves}
\begin{figure*}[t!]
    \begin{center}
    \includegraphics[width=\linewidth]{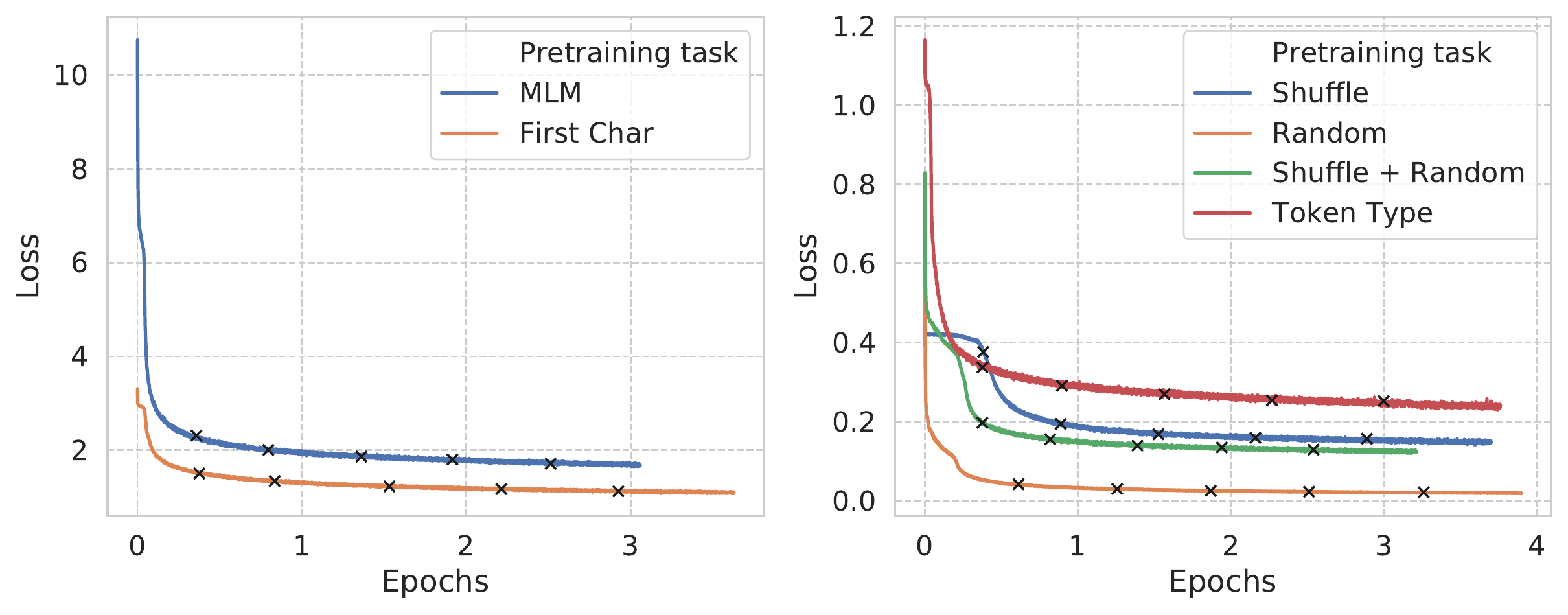}
    \caption{The loss curves for \bert{}-\textsc{base} models. Each $\times$ denotes a checkpoint pretrained for $1 \leq n \leq 5$ day(s).}
    \label{fig:base_loss}
    \end{center}
\end{figure*}

\begin{figure*}[t!]
    \begin{center}
    \includegraphics[width=\linewidth]{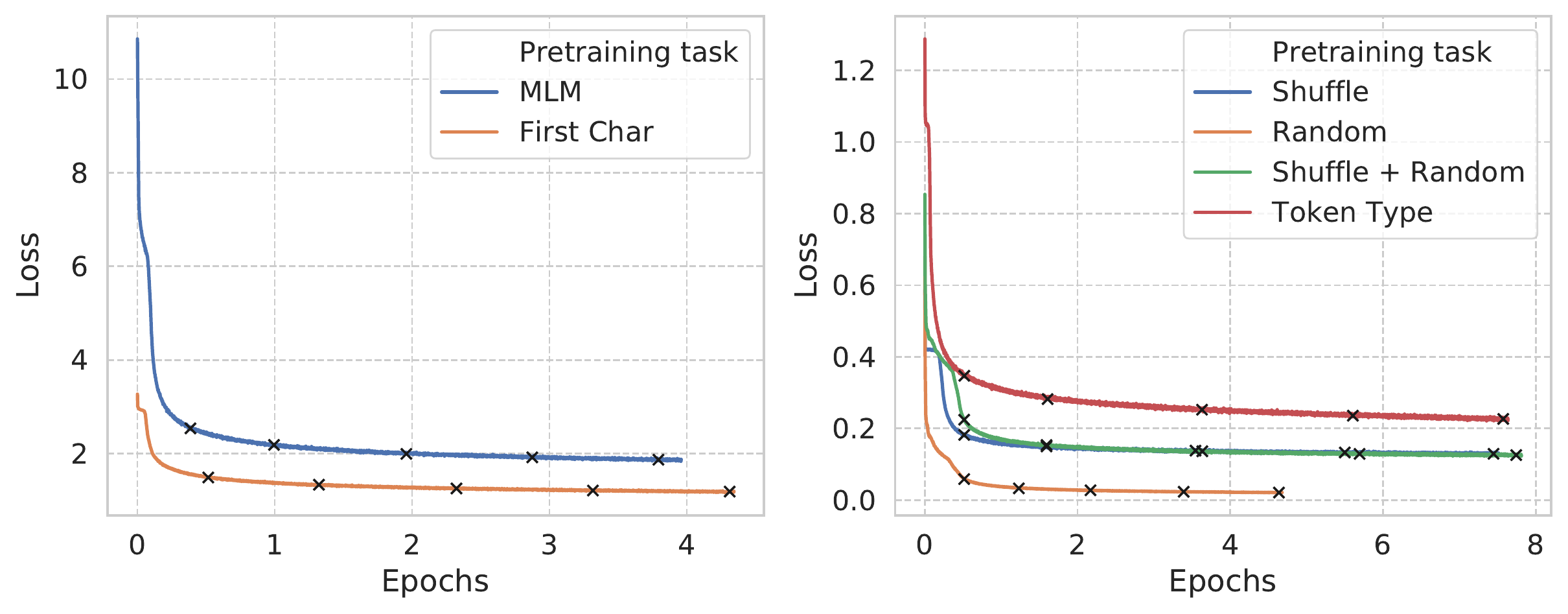}
    \caption{The loss curves for \bert{}-\textsc{medium} models. Each $\times$ denotes a checkpoint pretrained for $1 \leq n \leq 5$ day(s).}
    \label{fig:medium_loss}
    \end{center}
\end{figure*}

\begin{figure*}[t!]
    \begin{center}
    \includegraphics[width=\linewidth]{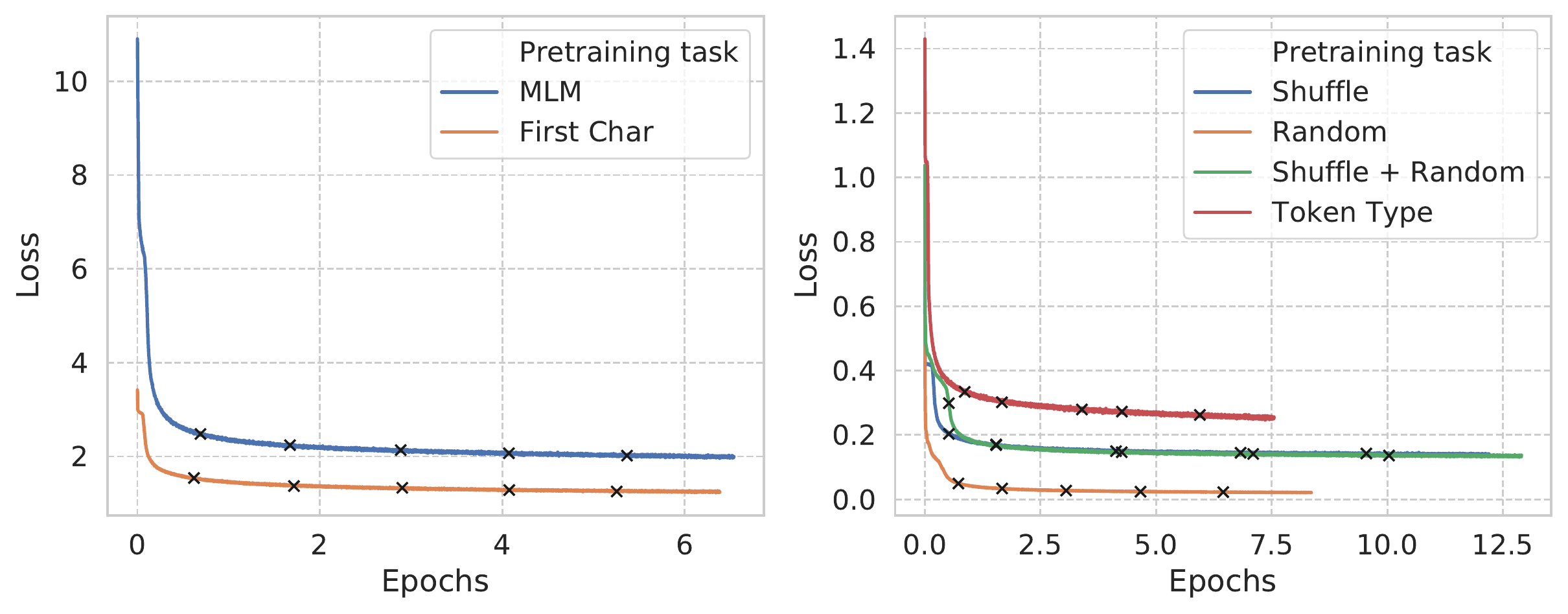}
    \caption{The loss curves for \bert{}-\textsc{small} models. Each $\times$ denotes a checkpoint pretrained for $1 \leq n \leq 5$ day(s).}
    \label{fig:small_loss}
    \end{center}
\end{figure*}

Figures \ref{fig:base_loss}, \ref{fig:medium_loss} and \ref{fig:small_loss} show the loss curves for our pretraining tasks in each model setting: \textsc{base}, \textsc{medium} and \textsc{small}. 

\begin{figure*}[t!]
    \begin{center}
    \includegraphics[width=\linewidth]{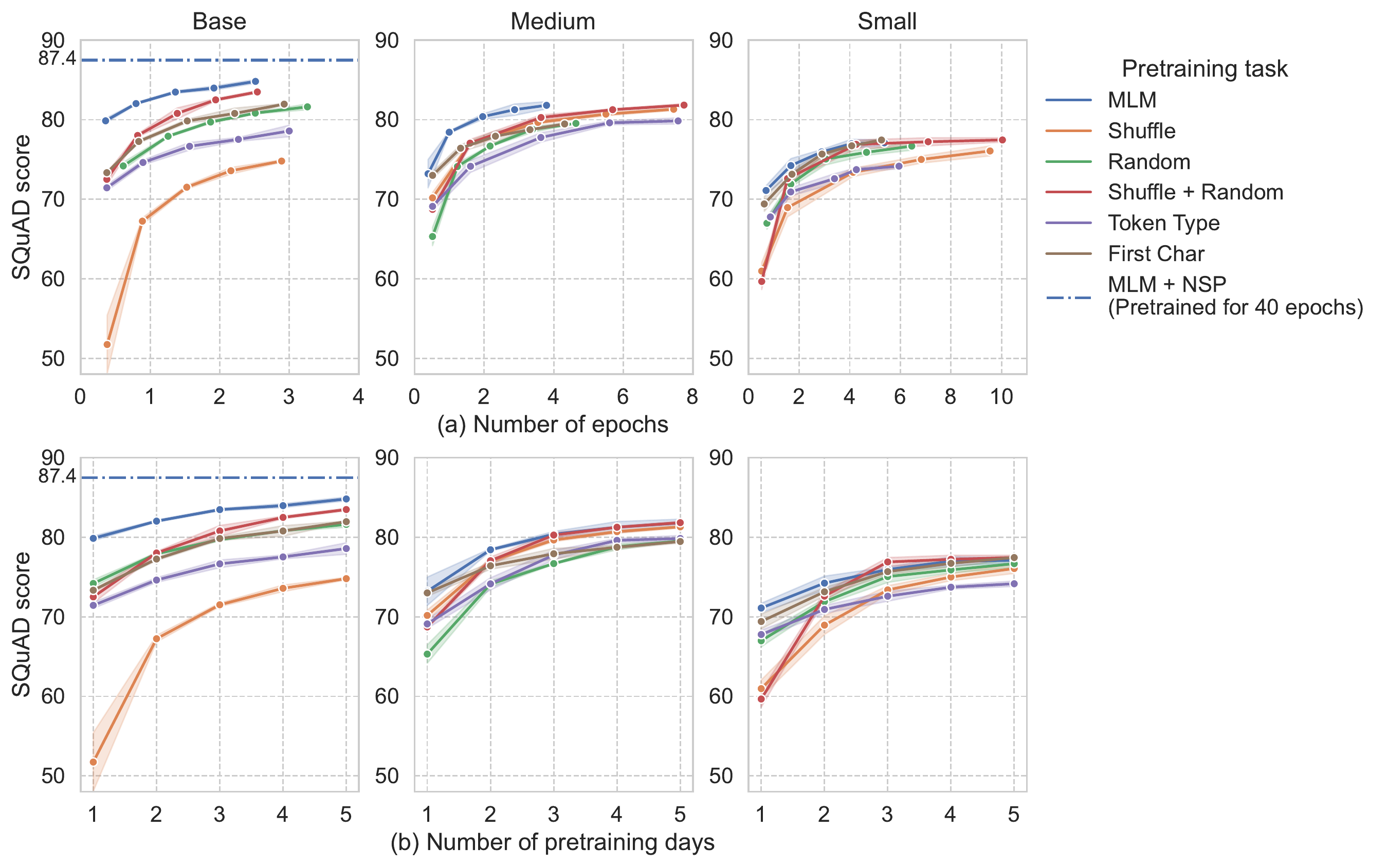}
    \caption{Results on \squad{} dev sets across (a) epochs and (b) days. Each point is a checkpoint pretrained for $1 \leq n \leq 5$ day(s).}
    \label{fig:ret_squad}
    \end{center}
\end{figure*}

\section{Performance in \squad{}}\label{sec:results_squad_appendix}

Figure \ref{fig:ret_squad} demonstrates the performance of our proposed methods across (a) epochs and (b) days in \squad{}.

\end{document}